\documentclass[conference]{IEEEtran}
\usepackage{url}
\usepackage{cite}

\usepackage{fancyhdr}

\usepackage{listings}
\usepackage{color}
\lstloadaspects{formats}

\definecolor{dkgreen}{rgb}{0,0.6,0}
\definecolor{gray}{rgb}{0.5,0.5,0.5}
\definecolor{mauve}{rgb}{0.58,0,0.82}

\ifCLASSINFOpdf
 \usepackage[pdftex]{graphicx}
 \graphicspath{{images/}}
\DeclareGraphicsExtensions{.pdf,.jpeg,.png}
\else
\fi
\begin{document}



%
\title{\LARGE Predictive Analysis of Chikungunya}

\author{\IEEEauthorblockN{Sayed Erfan Arefin\IEEEauthorrefmark{1},
Tasnia Ashrafi Heya\IEEEauthorrefmark{1} and Dr Moinul Zaber \IEEEauthorrefmark{1}}
\IEEEauthorblockA{Computer Science and Engineering Depertment,
BRAC University\\
66, Mohakhali, Dhaka\\
Email: \IEEEauthorrefmark{1}erfanjordison@gmail.com,
\IEEEauthorrefmark{1}tasnia.heya@gmail.com,
\IEEEauthorrefmark{1}zaber@du.ac.bd}}


%


\maketitle

\thispagestyle{fancy}

\begin{abstract}
Chikungunya   is an emerging threat for health security all over the world  which is spreading very fast. Researches for proper forecasting of the incidence rate of chikungunya has been going on in many places in which DARPA has done a very extensive summarized result from 2014 to 2017 with the data of suspected cases, confirmed cases, deaths, population and incidence rate in different countries. In this project, we have analysed the dataset from DARPA and extended it to predict the incidence rate using different features of weather like temperature, humidity, dewiness, wind and pressure  along with the latitude and longitude of every country. We had to use different APIs to find out these extra features from 2014-2016. After creating a pure dataset, we have used Linear Regression to predict the incidence rate and calculated the accuracy and error rate.
\end{abstract}



%
\IEEEpeerreviewmaketitle

\section{Introduction}
In December 2013, public health surveillance affirmed the main neighborhood transmission of Chikungunya infection in the Caribbean and only within three months, the infection spread from  Saint Martin island to six other neighboring islands including French Guiana and South America \cite{one}. Cassadou et al. and Omarjee et al. have elaborated the significance of proactive public health practice in such emergency situation of rising vector borne diseases \cite{two}. Most demonstrating papers center around the vector, yet, there is a detectable absence of research according to the the consideration of people into the numerical models. The vast majority of the work around this context is being done in France and Italy, which utilize diverse methodologies in their estimations [ECDC teleconference, January 2008] \cite{three}.

Schuler et al. has explained the epidemiological situation of TBE in Switzerland over a five year period, implying the heterogeneity of the frequency as indicated by cantons and the significance of the observation and inoculation as a preventive measure \cite{four}. To investigate the prediction of rising diseases like chikungunya, the Defense Advanced Research Projects Agency (DARPA) propelled the 2014– 2015  Chikungunya Challenge to conjecture the quantity of cases and spread of chikungunya illness all over America by comparing the incidence data reported to the Pan American Health Organization (PAHO) \cite{five}. Viable forecasting is doable with careful investigations but, there is a need to enhance the data quality to achieve more precise predicted results.

According to \cite{six}, data of the acute clinical profile, quality of life and consequent economic burden of the chikungunya patience was collected and summarized during the outbreak from May to September 2017, in Dhaka, Bangladesh. Analysing the quality of life of 1,326 chikungunya cases along with the analysis of major clinical variables have failed to show any statistically remarkable differences between confirmed and probable cases where their aim was to contribute for an effective syndromic surveillance system with early detection of future chikungunya outbreaks in resource-limited countries like Bangladesh.

\thispagestyle{empty}
\section{Data Collection and Filtering}

PAHO is the specialized international health agency for the Americas which also works with United Nations World Health Organization.
It works with countries throughout the region to improve and protect people's health. PAHO engages in technical cooperation with its member countries to fight communicable and noncommunicable diseases and their causes, to strengthen health systems, and to respond to emergencies and disasters.
We have collected reports from PAHO website on Chikungunya. PAHO’s data sources are: Data source: Cases reported by IHR NFPs to PAHO/WHO and/or through Member States websites.
In total we have collected 204 reports on Chikungunya from PAHO website. These were in pdf formats.
The reports contained the organization Names, logo, Data table and Notes. The data tables contained information of Chikungunya for the following parameters: Country/Region, Epidemiological Weeks, Suspected Cases, Confirmed Cases, Incident rates, Deaths, Imported Cases and Population.
 We converted these files to excel using https://pdftoxls.com 20 files at each batch conversion. After the conversion we selected all the tables from the converted files and saved only the tables. The files were converted to Microsoft Excel format (.xls) by the pdftoxls website.  Now we used the following command to convert the files to Open Spread Sheet format(.ods) in  order to import them to MySql Database using phpMyAdmin.

\begin{lstlisting}
// shell script
for i in *.xlsx; 
do 
	libreoffice --convert-to ods \$i; 
done

\end{lstlisting}

After the conversion to ods format we imported the files to MySql using phpMyAdmin. After that we filtered the data depending on some constrains. Some of the data tables of the reports didn’t had all the entries. Which may lead to a dataset with errors. We removed all the entries which didn’t had the following values: Epidemiological Weeks, Suspected Cases, Confirmed Cases. 
Due to conversion to excel from pdf, some of the Country names had garbage characters at the end. For example, some Country names contained \#,\^,g or \& at the end of the names, Epidemiological Weeks contained the word “Week”, “WEEk” with the number. We removed al the unnecessary characters from the end of the names. We used the following codes to filter the dataset. 

\begin{lstlisting}
$a = array( '>', '*', '(1)', '(2)', '(^)', '()', '\#', '\^', '?', '\$', '/', '\$', '\&' );
$filteredName = rtrim(rtrim(ltrim(str_replace( $a,"",\$row["Country"]))), 'g'); 
$week = rtrim(ltrim(str_replace( "WEEK","", $row["Epidemiological Weeks"]))); 
$week = rtrim(ltrim(str_replace( "Week","", $week))); 
$population = rtrim(ltrim(str_replace( ",","", $row["Population X 1000"])));


\end{lstlisting}

Then, we used the following SQL query “SELECT DISTINCT Country
FROM data” to get the distinct values of Countries from the Country column and saved them in a table. Later we used the Goolg egeocode api to get the latitude and longitude of the location. We wrote a php script that retrieved these values of the distinct Country names from the database and called the google geo code api for each entry in order to get the latitude and longitude of the places. The values were also saved in the database to the corresponding places. A google secret key was required to call the google geo code api, which was retrieved from the google cloud console by enabling google geo code api for the project. The block of code that did the mentioned job is given below.

\begin{lstlisting}
$sql = "SELECT DISTINCT Country FROM data";
$result = $conn->query($sql);
if ($result->num_rows > 0) {
     while($row = $result->fetch_assoc()) {
	$country =  $row['Country'] ;
	$service_url = 'https://maps.googleapis.com/maps/api/
	geocode/json?address='.urlencode
	 ($country ).'&key=<api key>';

	$curl = curl_init($service_url);
	curl_setopt($curl, CURLOPT_RETURNTRANSFER, true);
	$curl_response = curl_exec($curl);
	if ($curl_response === false) {
   		 $info = curl_getinfo($curl);
   		 curl_close($curl);
   		 die('error occured during curl exec. Additioanl info: ' . var_export($info));
	}
	curl_close($curl);
	\$decoded = json_decode($curl_response, true );
	if (isset($decoded->response->status) && $decoded->response->status == 'ERROR') {
 		die('error occured: ' . $decoded->response->errormessage);
	}
	$latitude = $decoded['results'][0]
		['geometry']['location']['lat'];
	$longitude = $decoded['results'][0]
		['geometry']['location']['lng'];
	$insertSql = 'INSERT INTO city_info VALUES( "'.$country.'", "'. $latitude .'","'. $longitude . '")';
	$result2 = $conn->query($insertSql);
     }
}

\end{lstlisting}

Now we have all the geo co ordinates of the Countries collected from the PAHO reports. 
Now we need to retrieve the weather information of the Countries of the reported Epidemiological Weeks. After searching for an API that will give us the weather information we chose “Dark Sky API” which provided weather information such as Temperature, Wind speed, Humidity, dew point, summary, Air pressure for first 1000 call for free for each day. Other APIs for getting historical weather data were not free, which included: OpenWeatherMap, Forcast.io, Weather Underground etc. 
The Dark Sky API requires a api key, latitude, longitude of the place and the unix time for which the weather data needs to be retrieved. 
We used the php DateTime() method to get the exact date of the 'Epidemiological Week of the given year.

\begin{lstlisting}
$year = $row['Year'];
$week = $row['Epidemiological Weeks'];
$id = $row['id'];

$dto = new DateTime();
$dto->setISODate($year, $week);
$startDate = $dto->format('Y-m-d');
$startDateTimeStamp = $dto->getTimestamp();

\end{lstlisting}

Later, we used the Google Timezone API to get the time zones of the countries using their geo coordinates and correct the dates which were converted from Epidemiological week. The code block that executed the mentioned process is given below.

\begin{lstlisting}
$latitude = $row['lat']; 
$longitude = $row['lon'];

//lets get the time zone
$service_url = 'https://maps.googleapis.com
/maps/api/timezone/json?
location='.$latitude.','.$longitude.'
&timestamp='.$startDateTimeStamp.'&key=<api key>';

$curl = curl_init($service_url);
curl_setopt($curl, CURLOPT_RETURNTRANSFER, true);
$curl_response = curl_exec($curl);
        if ($curl_response === false) {
            $info = curl_getinfo($curl);
            curl_close($curl);
            die('error occured during curl exec. Additioanl info: ' . var_export($info));
        }
        curl_close($curl);
        $decoded = json_decode($curl_response, true );
        if (isset($decoded->response->status) 
	&& $decoded->response->status == 'ERROR') {
            die('error occured: ' . $decoded->response->errormessage);
        }
         
$timeZoneId = $decoded['timeZoneId'];
$dto->setTimeZone(new DateTimeZone($timeZoneId));
$startDate = $dto->format('Y-m-d');
$startDateTimeStamp = $dto->getTimestamp();

\end{lstlisting}

Now we used this unix timestamp and geo coordinates to get the weather information for each entries. The php code that was used to get the weather information is as follows.

\begin{lstlisting}
$darksky = "https://api.darksky.net/forecast/
<api key>/".$latitude.",".$longitude.",
".$startDateTimeStamp;

$curl = curl_init($darksky);
curl_setopt($curl, CURLOPT_RETURNTRANSFER, true);
$curl_response = curl_exec($curl);
if ($curl_response === false) {
	$info = curl_getinfo($curl);
            curl_close($curl);
            die('error occured during curl exec. Additioanl info: ' . var_export($info));
 }
curl_close($curl);
$decoded = json_decode($curl_response, true );
if (isset($decoded->response->status) && $decoded->response->status == 'ERROR') {
        die('error occured: ' . $decoded->response->errormessage);
 }
         
$temperature =  $decoded['currently']['temperature'];
$summary = $decoded['currently']['summary'];
$dewPoint = $decoded['currently']['dewPoint'];
$humidity = $decoded['currently']['humidity'];
$pressure =  $decoded['currently']['pressure'];
$windSpeed = $decoded['currently']['windSpeed'];
$updateSql = "UPDATE data3 SET 
weather_windSpeed = '".$windSpeed."', 
weather_pressure = '".$pressure."', 
weather_temperature = '".$temperature."', 
weather_summary = '".$summary."', 
weather_dewPoint = '".$dewPoint."', 
weather_humidity = '".$humidity."' 
WHERE id=".$id;

$conn->query($updateSql);

\end{lstlisting}

Some entries such as, rows with country name Peru, Venezuela etc failed to get weather data from the Dark Sky API for the given date. We didn’t used those entries in our final dataset.

\section{Statistical Analysis}
\begin{figure}[ht!] 
 \centering
 \includegraphics[width=3.5in]{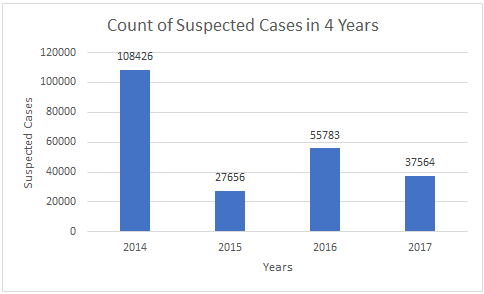}
 \caption{Count of Suspected Cases in 4 years}
 \label{1}
 \end{figure}

\begin{figure}[ht!] 
 \centering
 \includegraphics[width=3.5in]{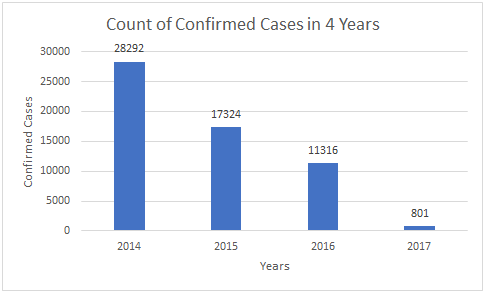}
 \caption{Count of Confirmed Cases in 4 years}
 \label{2}
 \end{figure}

\begin{figure}[ht!] 
 \centering
 \includegraphics[width=3.5in]{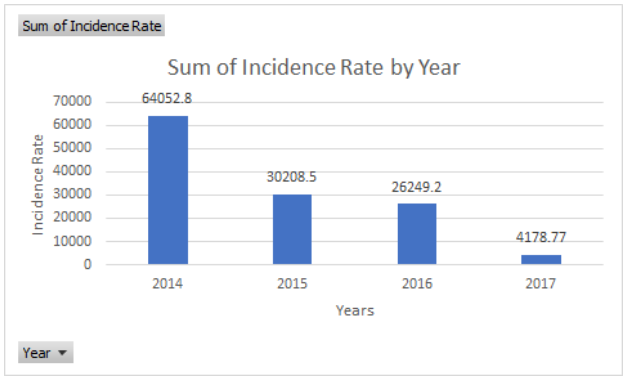}
 \caption{Sum of incidence rate by year}
 \label{3}
 \end{figure}

\begin{figure}[ht!] 
 \centering
 \includegraphics[width=3.5in]{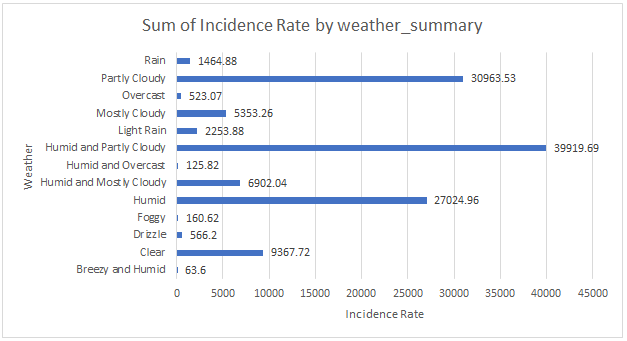}
 \caption{Count of incidence rate by weather\_summary}
 \label{4}
 \end{figure}

We can see from the Figure  ~\ref{1} and ~\ref{2} that, in 2014, both suspected and confirmed cases were higher than other three years which means the rate of spreading chikungunya virus was the highest in this year. On the other hand, confirmed cases were lowest in 2017 however, suspected cases were lowest in 2015. Yet, we can see from the incidence rate in Figure ~\ref{3}, that in after 2014, in 2015, rate of spreading this disease is higher than next two years by the total incidence rate.
From Figure  ~\ref{4} , we can see the effects of weather in spreading of chikungunya where it can be said that it is mostly spread in cloudy and humid weather from the graphical presentation from the dataset against total incidence rate .

\section{Result}
\begin{figure}[ht!] 
 \centering
 \includegraphics[width=3.5in]{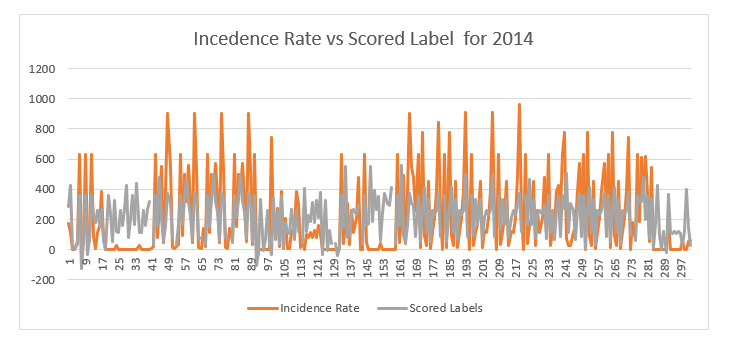}
 \caption{Comparison of scored results in 2014 }
 \label{compare_score_2014}
 \end{figure}

\begin{figure}[ht!] 
 \centering
 \includegraphics[width=3.5in]{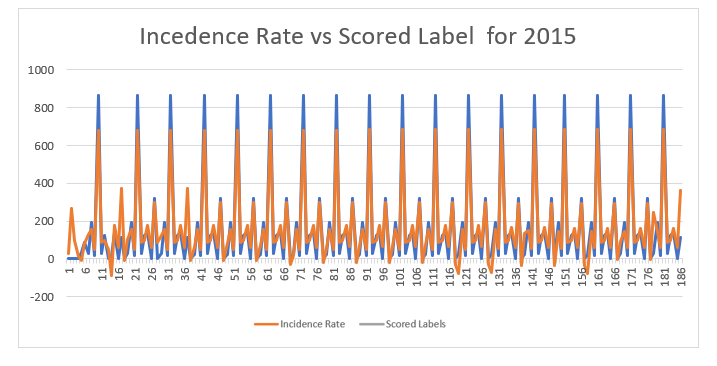}
 \caption{Comparison of scored results in 2015 }
 \label{compare_score_2015}
 \end{figure}

\begin{figure}[ht!] 
 \centering
 \includegraphics[width=3.5in]{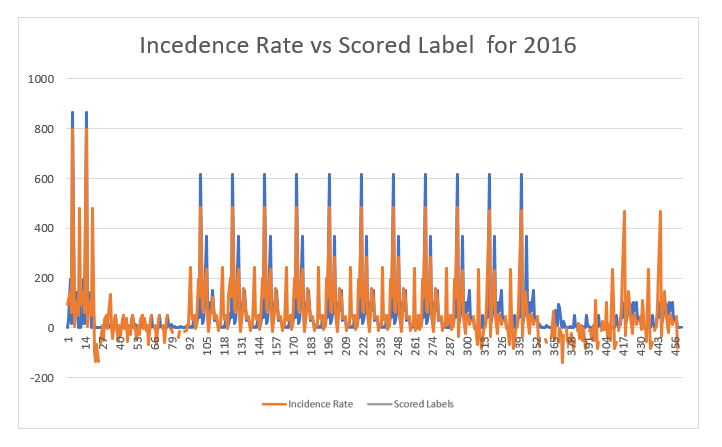}
 \caption{Comparison of scored results in 2016}
 \label{compare_score_2016}
 \end{figure}

\begin{figure}[ht!] 
 \centering
 \includegraphics[width=3.5in]{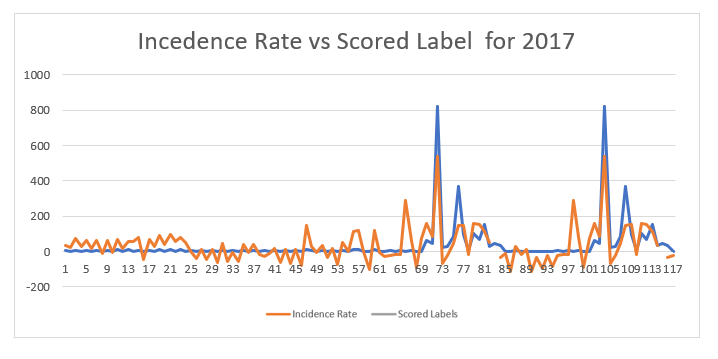}
 \caption{Comparison of scored results in 2017}
 \label{compare_score_2017}
 \end{figure}

 We have used Linear Regression for the Incidence Rate prediction on the extended and updated dataset considering Country, Week, Suspected, Confirmed, Imported cases, Deaths, Populatione X 1000, Year, Latitude, Longitude, weather\_temperature, weather\_summary, weather\_dewPoint, weather\_humidity, weather\_pressure and weather\_windSpeed as parameters. The comparison of the incidence rate and the scored label of the incidence rate by Linear regression for 2014, 2015, 2016 and 2017 is presented graphically in Fig ~\ref{compare_score_2014}, ~\ref{compare_score_2015}, ~\ref{compare_score_2016} and ~\ref{compare_score_2017} respectively. 

From Table~\ref{tab:table1}, we can see that the accuracy of our predicted incidence rate is 57.72\% having 43.2\% of Mean Absolute Error and 42.3\% of Relative Squared Error. The accuracy can be improved with more accurate dataset as there were a lot of missing data in the source dataset we have used. Those missing data could have consisted a great impact in the result with higher accuracy and less errors.

\begin{table}[h!]
  \begin{center}
    \caption{Result accuracy.}
    \label{tab:table1}
\begin{tabular}{|l|l|}
\hline
Mean Absolute Error          & 0.431976 \\ \hline
Relative Squared Error       & 0.422802 \\ \hline
Coefficient of Determination & 0.577198 \\ \hline
\end{tabular}
  \end{center}
\end{table}

\clearpage
\thispagestyle{empty}
\section{Conclusion}

By using the Microsoft Azure Machine learning Studio, Google GeoCode API, Google Timezone API and Dark Sky API, we were able to utilize the reports collected from PAHO on chikunguniya. We can use this trained models to predict the chances of Chikunguniya spread by providing weather information, geo coordinates and population.


\begin{thebibliography}{1}

\bibitem{one}
Semenza JC, Zeller H. Integrated surveillance for prevention and control of emerging vector-borne diseases in Europe. Euro Surveill. 2014;19(13):pii=20757. Available online: http://www.eurosurveillance.org/ViewArticle.aspx?ArticleId=20757

\bibitem{two}
Cassadou S, Boucau S, Petit-Sinturel M, Huc P, Leparc-Goffart I, Ledrans M. Emergence of chikungunya fever on the French side of Saint Martin island, October to December 2013. Euro Surveill. 2014;19(13):pii=20752.

\bibitem{three}
ECDC Meeting Report, Expert meeting on chikungunya modelling Stockholm, April 2008. Available online: https://ecdc.europa.eu/sites/portal/files/media/en/publications
/Publications/0804\_MER\_Chikungunya\_Modelling.pdf

\bibitem{four}
Schuler M, Zimmermann H, Altpeter E, Heininger U. Epidemiology of tick-borne encephalitis in Switzerland, 2005 to 2011. Euro Surveill. 2014;19(13):pii=20756.

\bibitem{five}
S Y. Del Valle, B H. McMahon, J Asher, R Hatchett, ‘Summary results of the 2014-2015 DARPA Chikungunya challenge’, 30 May 2018. Available at: [online] https://doi.org/10.1186/s12879-018-3124-7

\bibitem{six} 
M S Hossain , Md. M Hasan, M S Islam, S Islam, M Mozaffor, Md. A S Khan, ‘Chikungunya outbreak (2017) in Bangladesh: Clinical profile, economic impact and quality of life during the acute phase of the disease’, 2018 June 6. Available at: [online] 10.1371/journal.pntd.0006561

\bibitem{seven}
Chikungunya: Communicable Diseases and Health Analysis (CHA) - Communicable Diseases and Health Analysis (CHA), Data, Maps and Statistics. Available online: https://www.paho.org/hq/index.php?option=com\_topics\&
view=rdmore\&cid=5927\&item=chikungunya
\&type=statistics\&Itemid=40931\&lang=en


\end{thebibliography}
\end{document}